\documentclass{sig-alternate}

\usepackage[
  pass,
]{geometry}

\usepackage{multirow}
\usepackage{graphicx}
\usepackage{epstopdf}
\usepackage{url}

\usepackage{balance}
\usepackage{color}
\definecolor{Orange}{rgb}{0.9,0.5,0}
\definecolor{NavyBlue}{rgb}{0.1, 0.4, 0.8}
\definecolor{Magenta}{rgb}{0.8, 0.1, 0.6}
\definecolor{Green}{rgb}{0.1, 0.8, 0.3}
\definecolor{Brown}{rgb}{0.4, 0.3, 0.1}
\definecolor{Burgundy}{rgb}{0.5, 0.0, 0.13}
\definecolor{BrightCerulean}{rgb}{0.11, 0.67, 0.84}

\begin{document}

\CopyrightYear{2017} 
\setcopyright{acmcopyright} 
\conferenceinfo{EMDL'17,}{June 23, 2017, Niagara Falls, NY, USA}
\isbn{978-1-4503-4962-8/17/06}\acmPrice{\$15.00}
\doi{http://dx.doi.org/10.1145/3089801.3089802}

\title{Practical Processing of Mobile Sensor Data for Continual Deep Learning Predictions}

\numberofauthors{2}
\author{
\alignauthor
Kleomenis Katevas\thanks{Part of the work was done during a student internship at Telef\'onica Research.}\\
       \affaddr{Cognitive Science Research Group}\\
       \affaddr{Queen Mary University of London, UK}\\
       \email{k.katevas@qmul.ac.uk}
\alignauthor
Ilias Leontiadis, Martin Pielot, Joan~Serr{\`a}\\
       \affaddr{Telef{\'o}nica Research, Barcelona, Spain}\\
       \email{name.surname@telefonica.com}
}

\maketitle

\begin{abstract}

We present a practical approach for processing mobile sensor time series data for continual deep learning predictions. The approach comprises data cleaning, normalization, capping, time-based compression, and finally classification with a recurrent neural network. We demonstrate the effectiveness of the approach in a case study with 279~participants. On the basis of sparse sensor events, the network continually predicts whether the participants would attend to a notification within 10~minutes. Compared to a random baseline, the classifier achieves a 40\% performance increase (AUC of 0.702) on a withheld test set. This approach allows to forgo resource-intensive, domain-specific, error-prone feature engineering, which may drastically increase the applicability of machine learning to mobile phone sensor data.
\end{abstract}









\section{Background and Motivation}
\label{sec:background_motivation}

Machine learning can turn our mobile phones into sophisticated sensing and inference tools. Data captured from mobile phones cannot only be used to infer the location or level of acceleration of our phone, but also high-level information about, \emph{e.g.}, the environment, health \& well-being, and emotional states of the phone user~\cite{Lane:2010,Pielot:2015,Servia:2017}.

Traditional machine learning classifiers cannot typically handle raw sensor inputs, such as the level of activity as it is reported from the acceleration sensor. Therefore, sensor events have to be converted into features in order to become a relevant input for the classifier, such as the mean level of acceleration during a specified time window. Choosing which features to compute is an inherently time-consuming and creative task. Other than experience and domain knowledge, little guidelines exist on how to arrive to the best, or even to a sufficiently good set of features. Consequently, during feature extraction, important information may not be modeled and thus remain unused by the classifier. In addition, the extracted features may not be generic, in the sense that reusing them for a related but different task may be sub-optimal.

Deep learning proposes to solve this problem by learning the feature sets and the classifier at the same time, in a supervised way, and for a specific domain~\cite{Goodfellow16BOOK}. A cascade of neural network layers is employed, where each subsequent layer can learn more complex information, typically in a hierarchical fashion. The model implicitly identifies and learns predictive `features' from the available dataset and for the task at hand.

Deep learning is significantly outperforming state-of-the-art methods in several domains, such as image classification ~\cite{He15ICCV,Krizhevsky12NIPS}. A number of deep-learning architectures expect their input to be of a fixed size and format. However, in the context of mobile phone sensors, events which are predictive may be sparse and occur at irregular intervals. For example, in some use cases, the events of unlocking the screen or opening an app can be important predictors. These events, however, occur only rarely and asynchronously, making them hard to map into a fixed data format. Thus, there is no direct way to feed those events into a network that expects a stable-sized input. Recurrent neural networks (RNNs)~\cite{Goodfellow16BOOK} are more suited for variable-length sequential data, such as the one produced by mobile sensors. Nonetheless, they are typically designed for constant rate, synchronous sequences~\cite{Graves13ARXIV}. 

According to Lane \emph{et al.}~\cite{Lane:2015:DeepLearning}: ``\emph{If deep learning could lead to significantly more robust and efficient mobile sensor inference, it would revolutionize the field by rapidly expanding the number of sensor apps ready for mainstream usage}''. To achieve that, research is beginning to look into how deep learning models can deal with sparse and asynchronous sequences. For instance, Lee \emph{et~al.}~\cite{Lee:2016:PhasedLSTM} propose a phased-triggered RNN that uses a time gate to down-sample and discretize continuous sensor input, but is not capable of `de-sparsifying' sparse sensor data. 
DeepSense~\cite{Yao:2016:DeepSense} is, to our knowledge, the only work that inputs time series mobile sensor data into an RNN. In this work, the attendance to large time spans is achieved by using a combination of convolutional and RNN layers. The framework outperformed the baselines in several tasks (\emph{i.e.}, car tracking, activity recognition, and user identification). Even though the authors did not report any results with a wide range of mobile sensors, they claim that their framework can be directly applied to almost all other sensors, such as microphone, Wi-Fi signal, barometer, and light sensor.

In this paper, we propose a pipeline for the practical processing of sparse sensor data from mobile phones for the use in a deep learning classifier. Our goal is to enable continual predictions on the basis of sensor data streams, \emph{i.e.}, at each moment in time, the network should allow to make an estimation about the user's contextual state. The main points we tackle are: 
\begin{itemize} 
\item \emph{Data sparsity}. We propose a data format in which sparse sensor events are represented by positive numbers whereas absence of events is represented by zeros.

\item \emph{Temporal sparsity and asynchrony}. To improve performance, we propose a time-based compression method, which reduces the sparsity of the dataset.

\item \emph{User and class imbalance}. We consider and study four ground truth weighting strategies, used in the training of our deep learning models.
\end{itemize}

We demonstrate the effectiveness of our approach in a case study on a dataset of 279~mobile phone users, where sensor data and other events are used to continually predict whether the user will attend timely to a mobile phone notification. To the best of our knowledge, this is the first work that describes how to fuse a wide range of mobile sensors to predict the user's context using recurrent neural networks.
\section{Deep Learning Pipeline}
\label{sec:pipeline}

A deep neural network~\cite{Goodfellow16BOOK} is a series of fully connected layers of units (nodes) capable of mapping an input vector (raw data) into an output vector (\emph{e.g.}, inferred classes). A major difference with traditional machine learning is that instead of using manually crafted features as an input, deep networks are capable of using raw data (\emph{e.g.}, images, audio, text). An RNN is a specific type of deep network that takes sequential data as an input~\cite{Goodfellow16BOOK,Graves13ARXIV}. RNNs can be stateful, \emph{i.e.}, having an internal memory that allows them to remember past information. The most-used RNN architecture is the so-called long short-term memory (LSTM) network~\cite{LSTM}. It has repeatedly proven to be one of the best performing off-the-shelf approaches to sequence modeling.

\subsection{Prediction / Ground Truth}
RNNs typically learn from a continual series of events and ground truth labels. However, in the case of mobile sensor data, the ground truth labels can be sparse. For instance, in our case study, the ground truth is the comparably rare event of attending to a notification (only 1.45\% of the samples include ground truth labels). Thus, it is required that a prediction is happening continually, while the collected sensor data stream is being aggregated and the model is trained to do so, even in the absence of continual ground truth at the learning stage.

\subsection{Sensor Data Collection}
Mobile sensor data can be categorized into \emph{continuous}, where the sampling rate is fixed (\emph{e.g.}, accelerometer, light, \emph{etc.}) and \emph{event-driven}, where new data are reported when an event occurs (\emph{e.g.}, battery level drops, a notification is received, \emph{etc.}). When high precision from continuous sensors is not required, these data can be transformed into \emph{periodical}, by aggregating the data on custom time intervals (\emph{e.g.}, mean and maximum acceleration on every 10~minutes). Table~\ref{tab:sensors} describes the periodical and event-driven sensors of our case study.

\begin{table*}[!ht]
\small
\centering
\begin{tabular}{|c|l|p{13cm}|}
    \hline
    & Sensor             & Description \\
    \hline \hline
    \multirow{6}{*}[-3ex]{\rotatebox[origin=c]{90}{Periodical}}
    & Accelerometer      & Mean and maximum linear acceleration. \\
    & Battery            & Percentage of the device's battery drain per hour. \\
    & Data               & Network data activity in kb/sec (total received, total transmitted, cellular received, cellular transmitted). \\
    & Light              & Mean light level in lux. \\
    & Noise              & Mean noise levels in dB. \\
    & Semantic Location  & Location visited by the user, classified as \emph{Home}, \emph{Work}, \emph{Single} (visited once), \emph{Repeated} (visited regularly), \emph{Passing} (passed by for a short time), and \emph{Unknown}.\\

    \hline
    \multirow{9}{*}{\rotatebox[origin=c]{90}{Event-driven}}
    & App                & Name and category of the app that was opened by the user. \\
    & Audio Music        & Change in audio playback state (Music, No Music). \\
    & Audio Source       & Change in audio output of the device (Speaker, Headphones). \\
    & Charging State     & Charging state of the device (Charging, Not Charging). \\
    & Notification       & Post or removal of notification. \\
    & Notif. Center      & Event of accessing the device's notification center. \\
    & Ringer             & Change of the ringer mode (Normal, Silent, Vibrate). \\
    & Screen             & Change of the device's screen state (On, Off, Unlocked). \\
    & Screen Orientation & Change of screen orientation (Portrait, Landscape). \\
    \hline
\end{tabular}
\caption{Periodical and event-driven sensor data collected by a smartphone device.}
\label{tab:sensors}
\end{table*}

\subsection{Normalization and Capping}
Our input consists of real-valued sensor readings and one-hot encoded vectors. Before feeding the input into the network, we normalize it by re-scaling all the elements to lie between 0 and 1. 
In sensor data, we typically find highly-skewed, long-tail distributions. We empirically tested different thresholds above which the values are capped, and ended up using the $95^\text{th}$ percentile of the input data.

The time stamp of each data entry can be confusing for the RNN, as the value constantly increases over time (usually in epoch format, \emph{i.e.}, milliseconds since $1^\text{st}$ January 1970). Thus, we replace it with the \emph{time delta}, the time difference in minutes between the current and the previous sensor event. By capping the value at 60~minutes, we also avoid outliers in situations like the device is switched off for some time or the battery runs out.


\subsection{Fusing Sensors and Ground Truth}
RNNs are typically designed for synchronous data (\emph{e.g.}, audio, text, time series). While some of the sensors are sampled in regular intervals, most inputs in mobile data are event-driven. Therefore, they exhibit irregular bursts (see Table~\ref{tab:sensors} for some examples). Apart from introducing asynchronicity, event-driven inputs also result in extremely sparse input vectors.

\begin{figure}[t]
    \centering
    \includegraphics[width=0.95\columnwidth]{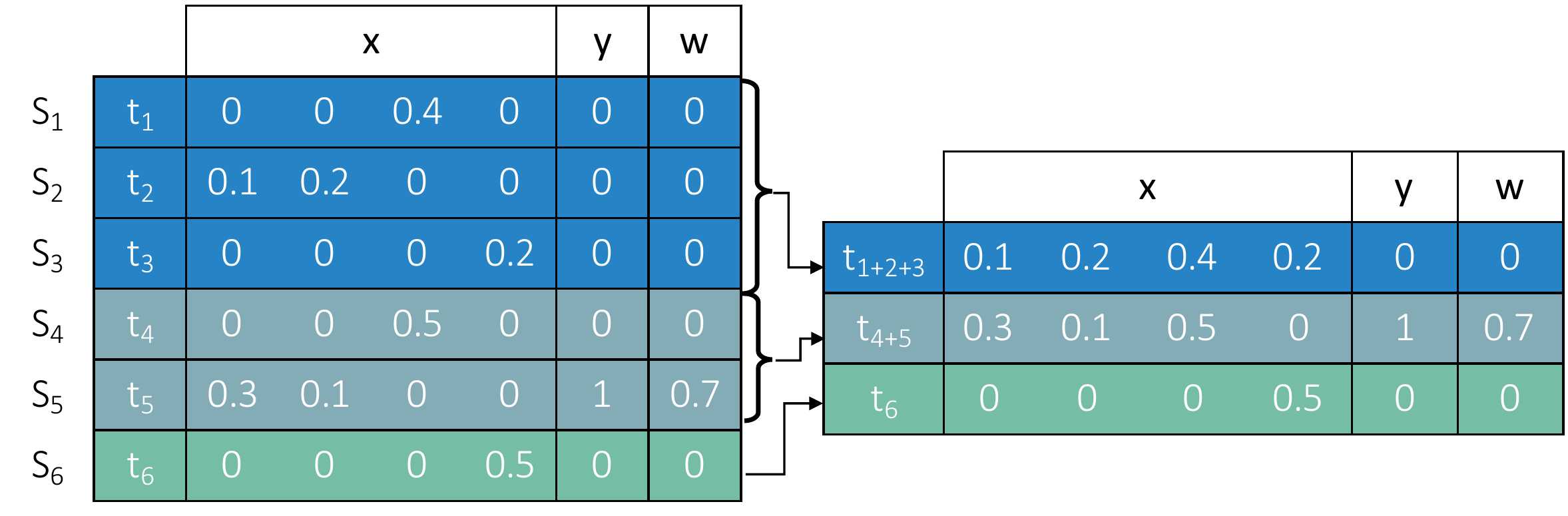}  
    \caption{Example of compressing sensor data.}
    \label{fig:compress}
\end{figure}

We organize the data in a form of sensor events, stored in a two dimensional matrix (Fig.~\ref{fig:compress}). Each row represents a sensor event ($S_i$), whereas each column represents a sensor measurement ($x$). Ground truth labels are also represented as a column in the matrix ($y$), using $w = 0$ when a ground truth label is not available. Since mobile phone sensors can be asynchronous and event-based, at every time step not all sensors can possibly provide data and ground truth labels. Therefore, we represent missing values with 0. To alleviate the issue of using 0 for both a missing value and a true 0 measurement, we re-scale data to range between 0.05 and 1.

\subsection{Structuring the Data for Training} 

The data is structured along several dimensions:

\noindent
\textbf{Input sample:} Each sample $i$ contains the input data of a single instance for a single user: a tuple $S_i = (x_i,y_i,w_i)$, where $x_i$ is a sensor data value, $y_i$ contains the ground truth label, and $w_i$ contains the weight of this sample, used in the error or loss function. Notice that not all samples contain a ground truth label (Fig.~\ref{fig:compress}). 

\noindent
\textbf{Sequences:} To train RNNs we need to provide for each user a time-ordered sequence of input samples. These samples are used to build an internal state that determines how past events affect future time slots. They are also used to back-propagate the error when training the RNN~\cite{Goodfellow16BOOK}. The number of steps to perform this back-propagation in time (\emph{sequence length}) is a parameter of the model.

\begin{figure*}[t]
    \centering
    \includegraphics[width=0.80\linewidth]{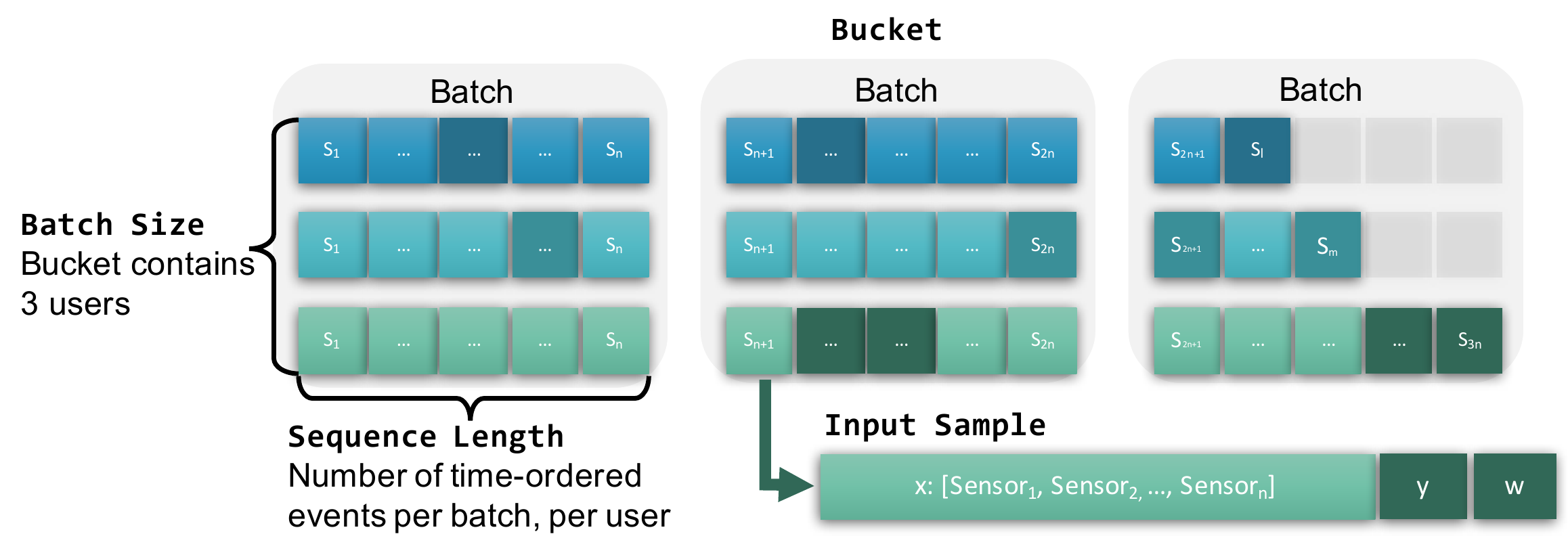}  
    \caption{Preparing data for training. Users are first split into buckets and then split into batches where multiple users are interleaved. Within each batch, a sequence of user data is provided. Some (but not all) of the samples contain the ground truth that will be used to train the model.}
\label{fig:batch}
\end{figure*}

\noindent
\textbf{Batches:} Modern deep learning techniques allow us to train a network in batches by interleaving multiple sequences together. Among others, batching allows to further exploit the power of matrix multiplication on the GPU and to avoid loading all data into memory at once. The \emph{batch size} has implications for the robustness of the error that is propagated in the learning phase~\cite{Keskar17ICLR}.
Figure~\ref{fig:batch} shows an example of 3~batches that encode 3~sequences of 5~samples each (15~samples per batch in total).

\noindent
\textbf{User buckets:} By using stateful RNNs, the internal RNN state is kept between two subsequent batches, potentially allowing it to learn sequences that are larger than the sequence length. To do so, we need to make sure that two subsequent batches interleave the same users with the same order. Therefore, we assign them into buckets: each bucket contains all the batches that are required to encode the data of its users. If the users within a bucket have a different number of sequences, we zero-pad their data and sort them so that the minimum zero padding is needed. Figure~\ref{fig:batch} shows an example of a single bucket that encodes the data of 3 users. 

\noindent
\textbf{Prediction:} The suggested arrangement into buckets and batches is only required in the training phase. For predictions we can even provide a single sample of a single user and the network will make a prediction based on the previous samples of that user.

\section{Performance Improvements}
\label{sec:perf_improvements}

\subsection{Time-Based Sparse Data Compression}

Batching with a single sensor event per sequence sample has two significant drawbacks: i) sub-optimal training where each event results in a training sample with very limited information contained in it, and ii) it is imposing a challenge to the RNN's internal states that now have to accommodate longer sequences to represent the same temporal context. Therefore, we perform an opportunistic, lossless compression of the input data: consecutive input samples are combined when there is no clashing information between them. The time delta for the merged samples is updated to indicate the overall elapsed time. More specifically, data from a subsequent sample $S_{i+1}$ can be merged into an existing sample $S_i$ only if all of the following rules are valid for all given input sensors $j$ (Fig.~\ref{fig:compress}):

\begin{itemize} 
\item  $S_{i}[j] = 0$ or $S_{i}[j] = S_{i+1}[j]$. In other words, we can only merge the next sample into the current one if the current value is zero (no existing data) or is equal to the value of the following sample.

\item $S_{i}[j]$ does not contain ground truth (sample weight is not zero).

\item The time delta between the merged samples is not larger than a threshold $T$ (we do not set $T$ in our experiments as our periodical sensors are configured to a fixed sampling rate of 10~minutes).
\end{itemize}


While this compression process results in a much denser input, there are some drawbacks. Firstly, a prediction is slightly delayed until a compressed sample has been generated. Smaller $T$ values can be used to shorten this delay. Secondly, the time information about the inter-arrival time of the compressed events is distorted. Finally, sensors that trigger multiple times with the same value can be compressed into a single event. However, performing a time-based compression presents a number of advantages that outweigh the previous drawbacks:

\begin{enumerate} 
\item Models train faster. With smaller sequences we have less samples to feed into the classifier. If those samples keep the same information (as it is the case), the process results in faster training times with no performance drop.

\item We have less elements in the sequence. This is important since the attention to past time spans of current RNN architectures is limited, a phenomenon known as the vanishing gradients problem~\cite{Pascanu12ICML}. Therefore, by compressing longer time spans into smaller sequences we can feed more information into the RNN. 

\item The sequence size is so small that we can even think of not deploying any further processing on the phone (including the deep network) and send that information to a server performing the remaining operations.
\end{enumerate}

\subsection{Sample weights}
\label{sec:weights}

Weights are traditionally used by machine-learning models to fight class imbalance: instances with significantly fewer samples typically get higher weights to force the model into considering them equally. In practice, the weights represent the contribution of each sample towards the loss function. However, in our case the weights are not only used to balance the different labels, but also for a more important task. 

As described, most of the generated samples simply contain sensor readings; there are very few samples that contain labeled data. Nevertheless, even if there are unlabeled sensor readings, all samples should go through the RNN as this will keep updating the internal RNN states. In other words, even if we don't want to make a prediction at time step $t$, this sample might affect a future time step $t+i$. An additional benefit is that by inputting every sequence, the network will make a prediction at every input and, in fact, we want to train the network like this. In the example of Figure~\ref{fig:batch}, we see that the whole sensor input is passed through the network but the network only learns from the highlighted samples.

Therefore, we need a way to indicate to the classifier that a given sample should be used to update the internal states (\emph{i.e.}, affect the past memory) but it should not be considered by the loss function in training time. To do so, we mark samples without a ground truth label with zeroed weights, whereas for instances that contain ground truth weights are calculated based on the number of instances of this label within each user. We consider 4~different strategies: i) no weights, therefore we resort to a simple binary indicator of whether to use the sample or not, ii) inverse frequency weighting~\cite{Manning08BOOK}, iii) inverse log-frequency weighting~\cite{Manning08BOOK}, and iv) inverse square root frequency weighting.


\section{Case study: Predicting Reactiveness to Notifications}
Notifications are alerts that try to attract the mobile phone user's attention to new content, such as unread emails or social network activity. While notifications help to avoid missing important content \cite{Pielot:2017}, they can have substantial negative effects. They disrupt and impair work performance, even when they are are not attended \cite{Stothart2015}. Constant exposure to notifications can negatively affect well-being \cite{Kushlev:2016}, as they induce symptoms of hyperactivity and inattention. At the same time, notifications are essential for people to keep up with expectations towards responsiveness \cite{Pielot:2017}. The research community is therefore investigating ways to reduce the negative effects of notifications. One approach that the community follows is to predict how reactive a user would be to a notification \cite{Turner:2015} to enable intelligent ways of handling them. In this section, we present a case study where we predict whether a user will react (click or dismiss it) to a mobile phone notification within a 10 minutes window. 


\subsection{Data collection}

Our dataset contains mobile phone use logs from\linebreak 279~Android phone users for an average duration of four weeks during summer 2016. The participants' ages ranged from 18 to 66 years ($M=37.7$, $SD=11.1$), with a balanced gender split (52.7\% female and 47.3\% male). The data was collected through an app which was running in the background while passively collecting rich sensor data about the user's context and phone usage. Participants registered the app to listen for notification and accessibility events, which allowed it to log what notifications participants received and after how much time they opened the corresponding app.


Table~\ref{tab:sensors} presents a list of all sensors used in this study. Based on the time stamp of each entry, we extracted some simple information such as the time delta (explained in Sec.~\ref{sec:pipeline}), the day of the week (1--7), the hour of the day (0--23), as well as a variable that indicates whether the current day is a working day or not (0--1). Basic demographics such as age and gender were also self-reported using a questionnaire at the beginning of the study and included in the dataset.

We computed the ground truth using 1 when a notification arrives and the user opens the app that originated the notification in less than 10 minutes, and 0 if the user either removed it from the notification center or just ignored it. We excluded all system and keyboard type notifications events from the ground truth, where a consequent action was not usually required by the user. The resulting dataset contains over 26~million phone usage events, about 1~million events after applying the sensor data compression, and about 388~thousand ground truth labels.

\subsection{Data Analysis}

For the analysis, we split the 4-week sequential dataset into training (first two weeks), validation ($3^\text{rd}$ week), known test ($4^\text{th}$ week) and unknown test ($4^\text{th}$ week). The difference between the two test sets is that the unknown test set includes 22~new users that the model has never seen before. Following the pipeline explained in Sec.~\ref{sec:pipeline}, we applied one-hot encoding to all categorical sensors, replaced all NaN values with zeros, applied normalization and capping, and finally applied time-based compression to the dataset.

To implement our model we used Keras v2.0.3~\cite{keras} with Theano v0.9~\cite{theano}. As an input layer we used a fully-connected time-distributed linear layer with 50 parametric rectified linear units~\cite{He15ICCV}. Two stateful LSTMs were used as hidden layers with 500 units. A final dense layer and a sigmoid activation function was applied to obtain output probabilities. We trained our model using standard cross-entropy loss~\cite{Goodfellow16BOOK} and the Adam optimizer~\cite{Kingma15ICLR} with default parameters.

As a baseline, we used a probability-based dummy classifier. On the basis of the training set, it determines, for each user, the probability that a notification of a certain category will be clicked within 10~minutes. In the prediction phase, it uses this probability as threshold in a random prediction. For example, if a user responded to 80\% of the WhatsApp messages within 10 minutes, the prediction will yield about 80\% positive predictions.

\balance

\subsection{Results}

\begin{figure}[t]
    \centering
    \includegraphics[width=0.95\columnwidth]{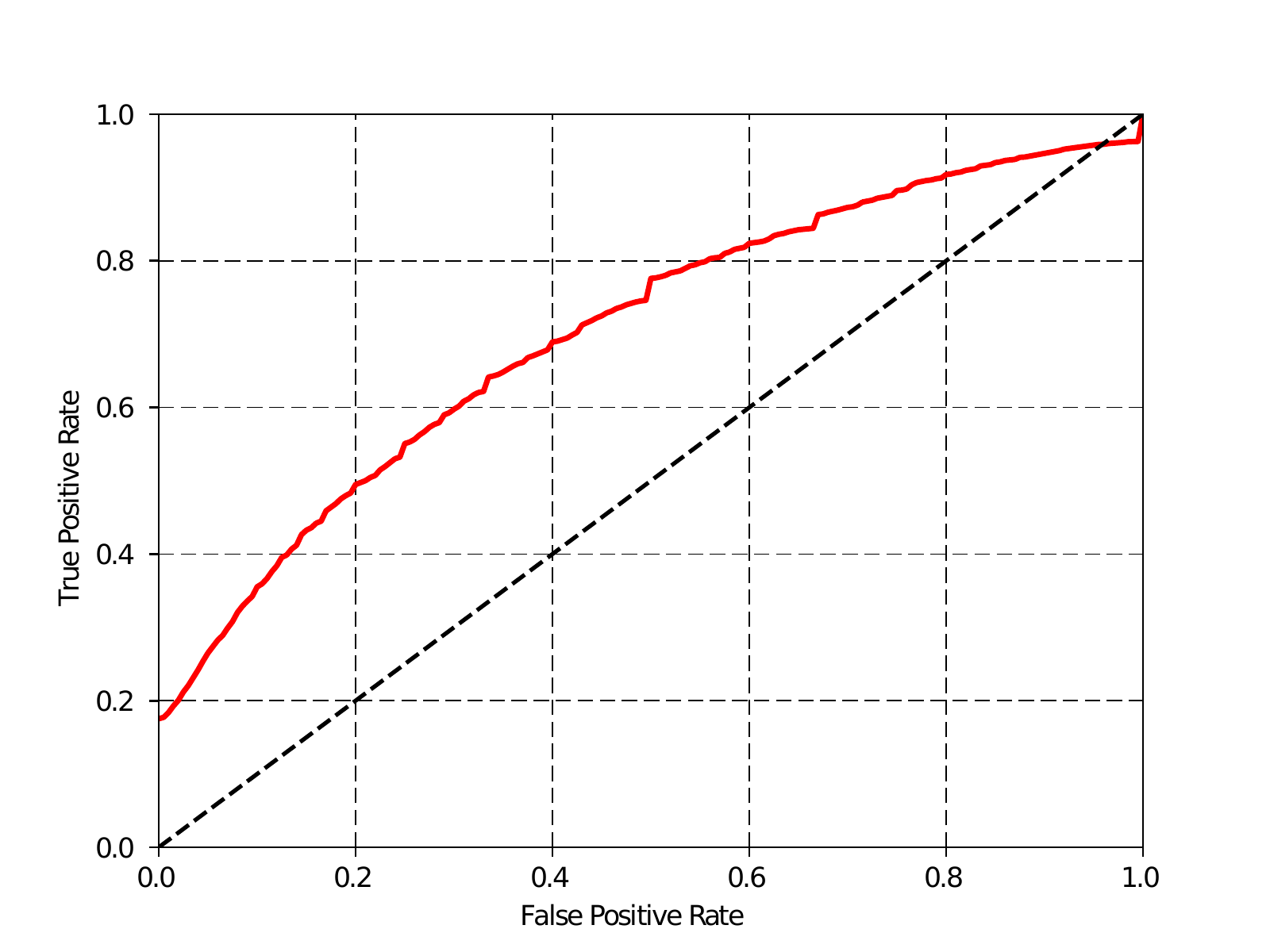} 
    \caption{ROC curve of the test set using the time-based compressed data and logarithmic weights (AUC = 0.702).}
    \label{fig:roc}
\end{figure}

In Table~\ref{tab:aucs}, we report the area under the curve (AUC) of the classifier using both the compressed and uncompressed datasets. The AUC is computed per user and per app category, and then averaged. Overall, we achieved an AUC of 0.70 in the test set and 0.69 in the unknown test set. Similar accuracies in both test sets suggest that the model is resilient to users outside the training set, which would be a very desirable property. By applying the time-based compression, we achieved a 95\% size reduction of the dataset and a 3.5\% relative improvement when predicting notification attendance. In addition, the model training time improved significantly from 1.3~hours to 2.8~minutes per epoch. We note that without the normalization and capping part described in Sec.~\ref{sec:pipeline}, the model presented some convergence issues. In Fig.~\ref{fig:roc} we report the performance of the classifier in a Receiver Operating Characteristic (ROC) curve.

In Table~\ref{tab:weights} we compare the accuracy of the model using the four considered types of weights (Sec.~\ref{sec:weights}). Apart from using the inverse of the frequency, which performed worse, we do not observe a substantial effect. Logarithmic weighting performed best in validation and test. Binary weights outperformed the rest in the case of the unknown test. However, due to its small size ($n=22$), the unknown test set was subject to high variance, preventing us from drawing clear conclusions.

\begin{table}[!ht]
\small
\centering
\begin{tabular}{|l|c|c|c|c}
    \hline 
                                & Valid         & Test           & Unknown Test  \\ \hline \hline
    Frequency                   & 0.679          & 0.681           & 0.669          \\ \hline
    Square root of frequency    & 0.697          & 0.698           & 0.667          \\ \hline
    No weights (binary)         & 0.706          & 0.700           & \textbf{0.696} \\ \hline
    Logarithm of frequency      & \textbf{0.713} & \textbf{0.702}  & 0.691         \\ \hline
\end{tabular}
\caption{AUC per weight type using the time-based compressed data.}
\label{tab:weights}
\end{table}

\begin{table}[!ht]
\small
\centering
\begin{tabular}{|l|c|c|c|c}
    \hline 
                    & Valid                & Test                   & Unknown Test          \\ \hline \hline
    Baseline        & 0.495  &  0.499   & 0.511   \\ \hline
    Uncompressed    & 0.688  &  0.678    & 0.671   \\ \hline
    Compressed      & \textbf{0.713}  &  \textbf{0.702}  & \textbf{0.691}   \\ \hline
\end{tabular}
\caption{AUCs using logarithmic weights.}
\label{tab:aucs}
\end{table}



\section{Conclusions and Future work}
\label{sec:conclusions}

We introduce a practical approach for preparing time series mobile sensor data for deep learning applications. We demonstrate its effectiveness in a case study with 279~participants. An RNN trained on data prepared with our approach achieved a 40\% performance increase with respect to a probabilistic random baseline in the task of predicting whether a notification would be attended within 10~minutes. We find that the model generalizes to unknown users without significant performance loss.

The proposed data processing approach enables running continual predictions on mobile sensor data streams. The proposed time-based compression further enables practical implementations, where the phone collects and compresses the data, and then sends it to server to run predictions. 
Future work includes the comparison of the performance to canonical approaches, the improvement of the compression strategy, and the potential application of more sophisticated deep learning techniques, such as transfer learning, or unsupervised learning with the use of generative adversarial networks.


\section{Acknowledgments}

The authors wish to thank the participants of the study. They also wish to thank Alexandros Karatzoglou for the useful discussions.

\bibliographystyle{abbrv}
\bibliography{references}

\begin{thebibliography}{10}

\bibitem{keras}
F.~Chollet.
\newblock Keras.
\newblock \url{https://github.com/fchollet/keras}, 2015.

\bibitem{Goodfellow16BOOK}
I.~Goodfellow, Y.~Bengio, and A.~Courville.
\newblock {\em {Deep learning}}.
\newblock MIT Press, Massachusetts, USA, 2016.

\bibitem{Graves13ARXIV}
A.~Graves.
\newblock {Generating sequences with recurrent neural networks}.
\newblock ArXiv: 1308.0850, 2013.

\bibitem{He15ICCV}
K.~He, X.~Zhang, S.~Ren, and J.~Sun.
\newblock {Delving deep into rectifiers: surpassing human-level performance on
  ImageNet classification}.
\newblock In {\em Proc. of the IEEE Int. Conf. on Computer Vision (ICCV)},
  pages 1026--1034, 2015.

\bibitem{LSTM}
S.~Hochreiter and J.~Schmidhuber.
\newblock Long short-term memory.
\newblock {\em Neural Comput.}, 9(8):1735--1780, Nov. 1997.

\bibitem{Keskar17ICLR}
N.~S. Keskar, D.~Mudigere, J.~Nocedal, M.~Smelyanskiy, and P.~T.~P. Tang.
\newblock {On large-batch training for deep learning: generalization gap and
  sharp minima}.
\newblock In {\em Proc. of the Int. Conf. on Learning Representations (ICLR)},
  2017.

\bibitem{Kingma15ICLR}
D.~P. Kingma and J.~L. Ba.
\newblock {Adam: a method for stochastic optimization}.
\newblock In {\em Proc. of the Int. Conf. on Learning Representations (ICLR)},
  2015.

\bibitem{Krizhevsky12NIPS}
A.~Krizhevsky, I.~Sutskever, and G.~Hinton.
\newblock {ImageNet classification with deep convolutional neural networks}.
\newblock In F.~Pereira, C.~J.~C. Burges, L.~Bottou, and K.~Q. Weinberger,
  editors, {\em Advances in Neural Information Processing Systems (NIPS)},
  volume~25, pages 1097--1105. Curran Associates Inc., 2012.

\bibitem{Kushlev:2016}
K.~Kushlev, J.~Proulx, and E.~W. Dunn.
\newblock "silence your phones": Smartphone notifications increase inattention
  and hyperactivity symptoms.
\newblock In {\em Proc CHI '16}, pages 1011--1020. ACM, 2016.

\bibitem{Lane:2015:DeepLearning}
N.~D. Lane and P.~Georgiev.
\newblock Can deep learning revolutionize mobile sensing?
\newblock In {\em Proceedings of the 16th International Workshop on Mobile
  Computing Systems and Applications}, HotMobile '15, pages 117--122. ACM,
  2015.

\bibitem{Lane:2010}
N.~D. Lane, E.~Miluzzo, H.~Lu, D.~Peebles, T.~Choudhury, and A.~T. Campbell.
\newblock A survey of mobile phone sensing.
\newblock {\em Comm. Mag.}, 48(9):140--150, Sept. 2010.

\bibitem{Manning08BOOK}
C.~D. Manning, P.~Raghavan, and H.~Sch{\"{u}}tze.
\newblock {\em {Introduction to information retrieval}}.
\newblock Cambridge University Press, Cambridge, UK, 2008.

\bibitem{Lee:2016:PhasedLSTM}
D.~Neil, M.~Pfeiffer, and S.-C. Liu.
\newblock Phased lstm: Accelerating recurrent network training for long or
  event-based sequences.
\newblock In D.~D. Lee, M.~Sugiyama, U.~V. Luxburg, I.~Guyon, and R.~Garnett,
  editors, {\em Advances in Neural Information Processing Systems (NIPS)},
  volume~29, pages 3882--3890. Curran Associates, Inc., 2016.

\bibitem{Pascanu12ICML}
R.~Pascanu, T.~Mikolov, and Y.~Bengio.
\newblock {On the difficulty of training recurrent neural networks}.
\newblock In {\em Proc. of the Int. Conf. on Machine Learning (ICML)}, pages
  1310--1318, 2013.

\bibitem{Pielot:2015}
M.~Pielot, T.~Dingler, J.~S. Pedro, and N.~Oliver.
\newblock When attention is not scarce - detecting boredom from mobile phone
  usage.
\newblock In {\em Proc. UbiComp '15}, UbiComp '15, pages 825--836. ACM, 2015.

\bibitem{Pielot:2017}
M.~Pielot and L.~Rello.
\newblock Productive, anxious, lonely - 24 hours without push notifications.
\newblock In {\em MobileHCI '17}, 2017.

\bibitem{Servia:2017}
S.~Servia-Rodr\'{\i}guez, K.~K. Rachuri, C.~Mascolo, P.~J. Rentfrow, N.~Lathia,
  and G.~M. Sandstrom.
\newblock Mobile sensing at the service of mental well-being: A large-scale
  longitudinal study.
\newblock In {\em Proc. WWW '17}, pages 103--112, 2017.

\bibitem{Stothart2015}
C.~Stothart, A.~Mitchum, and C.~Yehnert.
\newblock The attentional cost of receiving a cell phone notification.
\newblock {\em Journal of experimental psychology: human perception and
  performance}, 41(4):893, 2015.

\bibitem{theano}
{Theano Development Team}.
\newblock {Theano: A {Python} framework for fast computation of mathematical
  expressions}.
\newblock {\em arXiv e-prints}, abs/1605.02688, May 2016.

\bibitem{Turner:2015}
L.~D. Turner, S.~M. Allen, and R.~M. Whitaker.
\newblock Interruptibility prediction for ubiquitous systems: Conventions and
  new directions from a growing field.
\newblock In {\em Proc UbiComp '15}. ACM, 2015.

\bibitem{Yao:2016:DeepSense}
S.~Yao, S.~Hu, Y.~Zhao, A.~Zhang, and T.~Abdelzaher.
\newblock Deepsense: A unified deep learning framework for time-series mobile
  sensing data processing.
\newblock {\em arXiv preprint arXiv:1611.01942}, 2016.

\end{thebibliography}

\end{document}